\documentclass{article}

\PassOptionsToPackage{numbers,sort&compress}{natbib}
\usepackage[main, final]{neurips_2026}

\usepackage[utf8]{inputenc} 
\usepackage[T1]{fontenc}    
\usepackage{url}            
\usepackage{booktabs}       
\usepackage{amsfonts}       
\usepackage{nicefrac}       
\usepackage{microtype}      
\usepackage{xcolor}         
\usepackage{graphicx}
\usepackage{amsmath}
\usepackage{longtable}
\usepackage{bbm}
\definecolor{rliableolive}{HTML}{DCE6F0}
\usepackage[most]{tcolorbox}
\usepackage{wrapfig}
\usepackage{enumitem}
\usepackage{makecell}
\usepackage{tabularx} 
\usepackage{subcaption}
\usepackage{hyperref}       
\usepackage{pifont}          
\usepackage[table]{xcolor}

\newcommand{\cmark}{\textcolor{green!60!black}{\ding{51}}}
\newcommand{\xmark}{\textcolor{red!70!black}{\ding{55}}}
\tcbset{
  aibox/.style={
    width=\linewidth,
    top=4pt,
    bottom=4pt,
    colback=rliableolive!8!white,
    colframe=black,
    colbacktitle=black,
    enhanced,
    center,
    attach boxed title to top left={yshift=-0.1in,xshift=0.15in},
    boxed title style={boxrule=0pt,colframe=white,},
  }
}
\newcommand{\method}{\emph{v}OPD}
\newtcolorbox{AIbox}[2][]{aibox,title=#2,#1} 
\title{KL for a KL: On-Policy Distillation with Control Variate Baseline}
\author{%
  Minjae Oh,~Sangjun Song,~Gyubin Choi,~Yunho Choi,~Yohan Jo$^\dagger$ \\
  Seoul National University \\
  \texttt{\{kosair, yohan.jo\}@snu.ac.kr}
}

\begin{document}
\maketitle
\begin{abstract}
On-Policy Distillation (OPD) has emerged as a dominant post-training paradigm for Large Language Models, especially in reasoning domains. However, OPD remains unstable in practice due to the high gradient variance of its single-sample Monte Carlo estimator, and recipes for stable training are still immature. We propose \method{} (\textbf{O}n-\textbf{P}olicy \textbf{D}istillation with a control \emph{\textbf{v}}ariate baseline), which casts OPD as policy-gradient RL and stabilizes it by introducing a control variate baseline---canonically a value function---from the RL literature. We show that the OPD value function admits a closed form: the per-token negative reverse KL divergence between student and teacher, available directly from the existing forward pass with no additional critic or inference. Existing stabilization methods either compute the full token-level reverse KL over the entire vocabulary, adding significant overhead, or restrict it to a top-$k$ support, biasing the objective. \method{} instead preserves the lightweight single-sample estimator and subtracts the value function as a detached baseline to keep the gradient unbiased while reducing variance. Moreover, we show that a top-$k$ approximation of the baseline further lowers cost without compromising performance. Across mathematics, science, and instruction-following benchmarks, \method{} consistently outperforms base OPD and matches the most expensive full-vocabulary method, offering an efficient stabilization of OPD through principled RL variance reduction. Code: \href{https://github.com/holi-lab/vOPD}{\texttt{https://github.com/holi-lab/vOPD}}.
\end{abstract}

\def\thefootnote{\fnsymbol{footnote}}
\footnotetext[2]{Corresponding author.}
\def\thefootnote{\arabic{footnote}}

\section{Introduction}
Large Language Models (LLMs) have made remarkable advances in complex reasoning, accompanied by  improvements in post-training recipes~\citep{jaech2024openai, yang2025qwen3, team2025kimi}. A key driving factor has been Reinforcement Learning with Verifiable Rewards (RLVR)~\citep{lambert2024tulu, guo2025deepseek}, which trains LLMs directly against verifiable rewards---answer correctness, code execution outcomes---sidestepping the noise and reward hacking associated with learned reward models. RLVR has been successful thanks to its simple recipe, but this simplicity comes at a cost: LLMs generate thousands of intermediate tokens during reasoning before receiving a single scalar reward for the final answer. RLVR methods must therefore perform credit assignment over long chains of thought from this sparse signal. Such sparse supervision demands many rollouts and long training runs, making progress painfully slow.

On-Policy Distillation (OPD)~\citep{gu2024minillm, agarwal2024onpolicy} has emerged as an attractive alternative to RLVR when a strong teacher is available. Rather than relying on a sparse terminal reward, OPD minimizes the reverse KL divergence between the student and the teacher via dense, token-level signals, enabling faster training~\citep{lightman2024let}. Because it is on-policy and reward-driven, OPD can naturally be implemented using standard RL pipelines with a single-sample Monte Carlo estimator~\citep{lu2025onpolicydistillation}, and empirically matches RLVR accuracy at a fraction of the compute~\citep{patino2025opd}. Its effectiveness has been demonstrated in industrial-level post-training such as Qwen3, GLM-5, Nemotron-Cascade2, and DeepSeek-V4~\citep{yang2025qwen3, zeng2026glm, yang2026nemotron2, deepseekai2026deepseekv4}. Despite this success, OPD’s optimization recipe remains underdeveloped: training is often unstable in practice, and stabilization techniques are still immature relative to the recipes that drive successful RLVR training~\citep{yu2025dapo, khatri2025art, zheng2025gspo, chen2025minimax, liu2026tricks}. The most widely adopted fix replaces the single-sample estimator with a full-vocabulary token-level KL, incurring significant compute overhead~\citep{agarwal2024onpolicy}; a lighter-weight variant restricts the KL to a top-$k$ support, but this biases the gradient away from the true objective, still adds overhead, and yields only marginal gains~\citep{li2026rethinkingopd}. In contrast, we turn to the RL interpretation of OPD and propose a principled, low-cost method that reduces gradient variance while preserving the efficient single-sample Monte Carlo estimator.

We propose \textbf{\method{}} (\textbf{O}n-\textbf{P}olicy \textbf{D}istillation with a control \emph{\textbf{v}}ariate baseline), which leverages a standard tool from policy-gradient RL to reduce gradient variance: subtracting a control variate baseline~\citep{williams1992simple, sutton1999policygrad, greensmith2004variance}. Baseline subtraction underlies widely used methods, from actor-critic A3C and PPO to critic-free GRPO and RLOO~\citep{mnih2016asynchronous, schulman2017ppo, shao2024deepseekmath, ahmadian2024back} (see \S~\ref{sec:baseline}). Like these methods, \method{} reduces variance while keeping the gradient unbiased. The standard choice of baseline is the value function, and we show that for OPD this quantity admits a computable closed form: the per-token negative reverse KL divergence between the student and the teacher. The baseline is therefore available from the same forward pass that already computes the OPD objective---without an additional critic model, extra rollouts, or extra backward passes. We further show that the baseline can be approximated at lower cost using only the top-$k$ student tokens; crucially, because this approximation does not depend on the sampled token, it preserves the unbiasedness of the gradient regardless of $k$ (see \S~\ref{sec:vopd-topk}). Furthermore, we find empirically that the choice of $k$ has little effect on performance (see \S~\ref{sec:exp_ablation}).

We evaluate \method{} on seven diverse teacher--student pairs from the Qwen3~\citep{yang2025qwen3}, Olmo-3~\citep{olmo2025olmo}, and Gemma-4~\citep{team2026gemma} families across seven benchmarks spanning mathematics, science, and instruction following---MATH500~\citep{hendrycks2021math,lightman2024let}, Minerva Math~\citep{lewkowycz2022solving}, AMC23~\citep{maa_amc}, AIME24/25~\citep{maa_aime}, SciKnowEval~\citep{feng2024sciknoweval}, GPQA-Diamond~\citep{rein2024gpqa}, and IFEval~\citep{zhou2023instruction}---demonstrating consistent improvements over baseline methods. \method{} delivers absolute gains over base OPD of up to +3.6\% in average accuracy and +7.4\% on MATH500 (see \S\S~\ref{sec:exp-math}~and~\ref{sec:exp-sci}). Against the two stabilization variants, \method{} consistently outperforms top-$k$ OPD and matches full-vocabulary OPD while reducing wall-clock time by up to 57.7\%. We further validate the stability of \method{} through substantially lower gradient variance and selective damping of destabilizing reward tokens (see \S~\ref{sec:exp-further}). Overall, \method{} bridges RL and knowledge distillation, providing a principled, efficient approach to stable On-Policy Distillation.

\section{Preliminaries}
\label{sec:prelim}
\subsection{On-Policy Distillation}
\label{sec:opd}
On-Policy Distillation (OPD)~\citep{agarwal2024onpolicy,gu2024minillm} trains the student by minimizing the reverse KL divergence between the student ($\pi_\theta$) and the teacher ($\pi_T$):
\begin{equation}
\label{eq:RKL}
\mathbb{D}_\mathrm{KL}\!\left(\pi_\theta \,\|\, \pi_T\right)
\;=\;
\mathbb{E}_{x\sim \mathcal{D},\, y\sim \pi_\theta(\cdot\mid x)}
\!\left[\log \frac{\pi_\theta(y\mid x)}{\pi_T(y\mid x)}\right],
\end{equation}
where $x$ is a prompt drawn from a dataset $\mathcal{D}$ and $y = (y_1, \ldots, y_{|y|})$ is a response of length $|y|$. Importantly, OPD samples from the student during generation to obtain an unbiased estimator of the KL.  On-policy learning mitigates exposure bias~\citep{ranzato2015exposurebias}---the train-test discrepancy in off-policy training, where the model is trained on static data but conditions on its own outputs at test time. This has enabled effective training on long Chain-of-Thought reasoning tasks such as mathematics~\citep{wei2022chain,yang2025qwen3,lu2025onpolicydistillation}. In practice, Eq.~\eqref{eq:RKL} is commonly optimized via a single-sample Monte Carlo estimate, by maximizing the following token-level objective, where $c_t = (x, y_{<t})$ denotes the context at step $t$~\citep{lu2025onpolicydistillation}:
\begin{equation}
\label{eq:opd-objective}
\mathcal{J}_\text{OPD}(\theta)
\;=\;
\mathbb{E}_{x\sim\mathcal{D},\, y \sim \pi_\theta(\cdot\mid x)}
\!\left[\sum_{t=1}^{|y|} \log \frac{\pi_T(y_t \mid c_t)}{\pi_\theta(y_t \mid c_t)}\right].
\end{equation}
Following recent practice~\citep{lu2025onpolicydistillation, zeng2026glm, ko2026scaling}, Eq.~\eqref{eq:opd-objective} is optimized as policy-gradient 
RL~\citep{williams1992simple, sutton1999policygrad} by defining the per-token 
reward $r_t(c_t, y_t) = \log \pi_T(y_t \mid c_t) - \log \pi_\theta(y_t \mid c_t)$  as a fixed scalar with no gradient flowing through it. This yields the following policy gradient:
\begin{equation}
\label{eq:opd-gradient}
\nabla_\theta \mathcal{J}_\text{OPD}(\theta)
\;=\;
\mathbb{E}_{x\sim\mathcal{D},\, y\sim \pi_\theta(\cdot\mid x)}
\!\left[\sum_{t=1}^{|y|}
\underbrace{\bigl(\log \pi_T(y_t \mid c_t) - \log \pi_\theta(y_t \mid c_t)\bigr)}_{\textstyle r_t(c_t,y_t)}
\,\nabla_\theta \log \pi_\theta(y_t \mid c_t)\right].
\end{equation}
We refer to this base formulation as \textbf{OPD} throughout the paper. Its backward pass touches only $\nabla_\theta\log \pi_\theta(y_t\mid c_t)$ at the single sampled token, making it the most computationally efficient variant. However, the single-sample Monte Carlo estimator carries high variance, leading to training instability. We next discuss two variants that aim to stabilize OPD, along with the drawbacks of each.

\paragraph{Full-vocabulary OPD ($\text{OPD}_\text{full-V}$).}
To mitigate the variance of the single-sample estimator, one variant
computes the full per-token KL over the entire vocabulary $\mathcal{V}$~\citep{agarwal2024onpolicy}:
\begin{equation}
\label{eq:opd-full-J}
\mathcal{J}_{\text{OPD}_\text{full-V}}(\theta)
\;=\;
-\,\mathbb{E}\!\left[\, \sum_{t=1}^{|y|} \mathbb{D}_\mathrm{KL}\!\bigl(\pi_\theta(\cdot\mid c_t) \,\big\|\, \pi_T(\cdot\mid c_t)\bigr) \,\right]
\;=\;
\mathbb{E}\!\left[\, \sum_{t=1}^{|y|} \sum_{v \in \mathcal{V}} \pi_\theta(v\mid c_t)\,r_t(c_t,v) \,\right],
\end{equation}
where $r_t(c_t,v) = \log\pi_T(v\mid c_t) - \log\pi_\theta(v\mid c_t)$ extends
the per-token reward to any vocabulary entry. Similar to Eq.~\eqref{eq:opd-gradient}, Eq.~\eqref{eq:opd-full-J} can be optimized by the corresponding gradient:
\begin{equation}
\label{eq:opd-full-grad}
\nabla_\theta\mathcal{J}_{\text{OPD}_\text{full-V}}(\theta)
\;=\;
\mathbb{E}\!\left[\, \sum_{t=1}^{|y|} \sum_{v \in \mathcal{V}} \pi_\theta(v\mid c_t)\,r_t(c_t,v)\,\nabla_\theta\log \pi_\theta(v\mid c_t) \,\right],
\end{equation}
which is the exact expectation of Eq.~\eqref{eq:opd-gradient} under $v \sim \pi_\theta$, and is therefore zero-variance with respect to token sampling for a given $c_t$. The cost, however, is substantial, as it requires a backward pass through the full vocabulary at every token position (e.g., $|\mathcal{V}| \approx 150\mathrm{k}$ for Qwen3~\citep{yang2025qwen3}).

\paragraph{Top-$k$ OPD ($\text{OPD}_\text{top-$k$}$).}
A lightweight variant of $\text{OPD}_\text{full-V}$ computes the per-token KL against only the top-$k$ tokens, with $k \ll |\mathcal{V}|$~\citep{li2026rethinkingopd}. We consider the student-side top-$k$ version, which restricts the KL to the support $S_t$ of the student's $k$ most likely tokens:
\begin{equation}
\label{eq:opd-topk-J}
\mathcal{J}_{\text{OPD}_\text{top-$k$}}(\theta)
\;=\;
-\,\mathbb{E}\!\left[\, \sum_{t=1}^{|y|} \mathbb{D}_\mathrm{KL}\!\bigl(\bar\pi_\theta(\cdot\mid c_t) \,\big\|\, \bar\pi_T(\cdot\mid c_t)\bigr) \,\right],
\quad
\bar\pi(v\mid c_t) = \frac{\pi(v\mid c_t)\,\mathbf{1}[v \in S_t]}{\sum_{u \in S_t} \pi(u\mid c_t)}.
\end{equation}
Analogously, Eq.~\eqref{eq:opd-topk-J} is optimized by a gradient of the same shape as Eq.~\eqref{eq:opd-full-grad}, but restricted to the support $S_t$ and acting on the renormalized distributions:
\begin{equation}
\label{eq:opd-topk-grad}
\nabla_\theta\mathcal{J}_{\text{OPD}_\text{top-$k$}}(\theta)
\;=\;
\mathbb{E}\!\left[\, \sum_{t=1}^{|y|} \sum_{v \in S_t}
\bar\pi_\theta(v\mid c_t)\,\log\frac{\bar\pi_T(v\mid c_t)}{\bar\pi_\theta(v\mid c_t)}
\nabla_\theta\log \bar\pi_\theta(v\mid c_t) \,\right].
\end{equation}
The backward pass now flows through $k$ tokens per position, rather than the single sampled token in OPD---substantially more lightweight than the full vocabulary, but heavier than base OPD. More importantly, this comes at the cost of \emph{bias}: $\nabla_\theta\mathcal{J}_{\text{OPD}_\text{top-$k$}} \neq \nabla_\theta\mathcal{J}_\text{OPD}$, since restricting to $S_t$ omits out-of-support mass. In practice, despite this added compute, $\text{OPD}_\text{top-$k$}$ has been reported to yield only marginal gains over base OPD~\citep{li2026rethinkingopd}, an observation we confirm in \S~\ref{sec:exp-math}.

\subsection{Control Variate Baseline in Reinforcement Learning}
\label{sec:baseline}
As OPD in Eq.~\eqref{eq:opd-gradient} is a form of RL, we now introduce the standard variance-reduction tool used in policy-gradient RL: subtracting a baseline $b_t(c_t)$ from the per-step reward, yielding the \emph{advantage} $a_t(c_t,y_t) = r_t(c_t,y_t) - b_t(c_t)$~\citep{williams1992simple,sutton1999policygrad}, and the corresponding gradient:
\begin{equation}
\label{eq:baseline-grad}
\nabla_\theta \mathcal{J}(\theta)
\;=\;
\mathbb{E}\!\left[\, \sum_{t=1}^{|y|} \bigl(r_t(c_t,y_t) - b_t(c_t)\bigr)\,\nabla_\theta\log \pi_\theta(y_t\mid c_t) \,\right].
\end{equation}
This has two properties that make it successful in modern RL. \textbf{(i)~Unbiasedness:} for \emph{any} $b_t$ that is independent of the sampled action $y_t$, we have $\mathbb{E}[b_t(c_t)\nabla_\theta\log\pi_\theta(y_t\mid c_t)] = 0$, so the expected gradient is unchanged and the estimator remains unbiased (see \S~\ref{app:unbiased}). \textbf{(ii)~Variance reduction:} a well-chosen baseline reduces the gradient variance, the canonical choice being the value function $V^{\pi_\theta}(c_t) = \mathbb{E}_{y_t \sim \pi_\theta(\cdot\mid c_t)}[r_t(c_t,y_t)]$ (see \S~\ref{app:variance}). Baseline subtraction underlies the success of essentially every modern policy-gradient algorithm, from classical actor-critic methods with a learned value baseline~\citep{mnih2016asynchronous, schulman2017ppo,schulman2016high} to the group-relative baseline in GRPO~\citep{shao2024deepseekmath}.

\section{Control Variate Baseline for OPD}
\label{sec:method}
We introduce \method{} (\textbf{O}n-\textbf{P}olicy \textbf{D}istillation with a control \emph{\textbf{v}}ariate baseline), which addresses the high variance of OPD (see \S~\ref{sec:opd}) by exploiting its RL interpretation and subtracting a control variate baseline (see \S~\ref{sec:baseline}). \method{} is an unbiased, lower-variance version of OPD that requires no additional backward passes, making it computationally efficient. We first show that the value function of OPD is available in closed form as the per-step reverse KL, and discuss the loss formulation of \method{} (see \S~\ref{sec:vopd_value}). We then propose an even more computationally efficient version using a top-$k$ KL estimate (see \S~\ref{sec:vopd-topk}), and compare our methods with the various variants of OPD (see \S~\ref{sec:opd-comparison}).

\subsection{The Value Function of OPD}
\label{sec:vopd_value}
As discussed in \S~\ref{sec:baseline}, the standard choice of baseline is the value function. Recall the OPD per-token reward $r_t(c_t, y_t) = \log \pi_T(y_t\mid c_t) - \log \pi_\theta(y_t\mid c_t)$ from Eq.~\eqref{eq:opd-gradient}. By definition, taking the expectation of $r_t$ under the student distribution ($\pi_\theta$) gives the per-step value function:
\begin{equation}
\label{eq:value-fn}
V^{\pi_\theta}(c_t)
\;=\;
\mathbb{E}_{y_t \sim \pi_\theta(\cdot\mid c_t)}[r_t(c_t, y_t)]
\;=\;
-\,\mathbb{D}_\mathrm{KL}\!\bigl(\pi_\theta(\cdot\mid c_t) \,\big\|\, \pi_T(\cdot\mid c_t)\bigr).
\end{equation}
The value function is thus exactly the negative per-step reverse KL~\citep{tang2025few}, computable in closed form using the already-computed student ($\pi_\theta$) and teacher ($\pi_T$) distributions at context $c_t$, without a learned value network, extra rollouts, or additional backward passes. Substituting Eq.~\eqref{eq:value-fn} as the baseline in Eq.~\eqref{eq:baseline-grad} gives the \method{} gradient estimator with advantage $a_t(c_t, y_t)$:
\begin{equation}
\label{eq:vopd-full-grad}
\nabla_\theta \mathcal{J}_{\text{\method{}}}(\theta)
\;=\;
\mathbb{E}\!\left[\, \sum_{t=1}^{|y|}
\underbrace{\Bigl( r_t(c_t, y_t) + \mathbb{D}_\mathrm{KL}\!\bigl(\pi_\theta(\cdot\mid c_t) \,\big\|\, \pi_T(\cdot\mid c_t)\bigr) \Bigr)}_{\textstyle a_t(c_t,y_t)}
\,\nabla_\theta \log \pi_\theta(y_t\mid c_t) \,\right],
\end{equation}
which is denoted by \method$_{\text{full-V}}$; the subscript indicates the expectation over the full vocabulary to compute the KL baseline. As discussed in \S~\ref{sec:baseline}, this estimator has the same expected gradient as OPD: $\nabla \mathcal{J}_{\text{\method}_{\text{full-V}}} = \nabla \mathcal{J}_\text{OPD}$. Importantly, the baseline KL is computed only in the forward pass and does not propagate gradients through the vocabulary, so the backward pass flows only through $\nabla_\theta \log \pi_\theta(y_t\mid c_t)$ at the single sampled token, identical to base OPD.

\paragraph{Variance reduction.}
We now examine where and why \method{} reduces variance, showing it dampens gradients most strongly on the most destabilizing cases. Recent works have identified \emph{high-mismatch tokens}---where the student and the teacher distributions strongly disagree---as the dominant source of OPD's gradient instability: at these tokens, the per-token reward ($r_t$) takes large negative values, producing heavy-tailed gradients that dominate training~\citep{ko2026scaling, li2026rethinkingopd}. \method{}'s baseline directly counteracts this. Since $-V^{\pi_\theta}(c_t) = \mathbb{D}_\mathrm{KL}(\pi_\theta\|\pi_T)$ becomes a large positive value precisely when the student and the teacher strongly disagree, the \method{} advantage $(a_t=r_t+\mathbb{D}_\mathrm{KL}(\pi_\theta\|\pi_T))$ stays bounded even on such heavy-tailed tokens, acting as a regularizer. This token-level reward damping translates directly into lower gradient variance, and we show that the per-token reduction of \method{} is approximately:
\begin{equation}
\label{eq:variance-reduction}
\underbrace{\mathrm{tr}\!\bigl(\mathrm{Var}[g_\text{OPD}]\bigr)}_{\text{OPD variance}}
\;-\;
\underbrace{\mathrm{tr}\!\bigl(\mathrm{Var}[g_{\text{\method}_{\text{full-V}}}]\bigr)}_{\text{\method{} variance}}
\;\approx\;
\mathbb{D}_\mathrm{KL}\!\bigl(\pi_\theta(\cdot\mid c_t) \,\big\|\, \pi_T(\cdot\mid c_t)\bigr)^{\!2}
\;\cdot\;
\mathbb{E}_{\pi_\theta}\!\bigl[\,\|\nabla_\theta \log \pi_\theta(y_t\mid c_t)\|^2\,\bigr],
\end{equation}
where $g_\text{OPD}$ and $g_{\text{\method}_{\text{full-V}}}$ are the gradient estimators of OPD (Eq.~\eqref{eq:opd-gradient}) and \method$_{\text{full-V}}$ (Eq.~\eqref{eq:vopd-full-grad}), and $\mathrm{tr}(\cdot)$ denotes the matrix trace. We provide a detailed derivation in \S~\ref{app:variance-derivation}. From Eq.~\eqref{eq:variance-reduction}, the variance reduction is largest when the squared $\mathbb{D}_\mathrm{KL}(\pi_\theta\|\pi_T)$ is large at the high-mismatch tokens, which matches our token-level reward damping view. Overall, \method{} dampens the noisy negative long-tail gradients destabilizing OPD, which we further validate empirically in \S~\ref{sec:exp-further}.

\paragraph{Connection to $\text{OPD}_\text{full-V}$.}
A natural question is whether the same baseline could also help $\text{OPD}_\text{full-V}$. The answer is no, and the reason illuminates the relationship between the two methods. Subtracting the value baseline from the $\text{OPD}_\text{full-V}$ gradient (Eq.~\eqref{eq:opd-full-grad}) gives:
\begin{equation}
\label{eq:full-v-with-baseline}
\nabla_\theta \mathcal{J}_{\text{OPD}_\text{full-V}}^{\text{+baseline}}
\;=\;
\mathbb{E}\!\left[\, \sum_{t=1}^{|y|} \sum_{v \in \mathcal{V}}
\pi_\theta(v\mid c_t)\,
\underbrace{\bigl(r_t(c_t,v) - V^{\pi_\theta}(c_t)\bigr)}_{\textstyle a_t(c_t,v)}
\,\nabla_\theta\log \pi_\theta(v\mid c_t) \,\right],
\end{equation}
which is identical to the original gradient because the baseline contribution vanishes:
\begin{equation}
\label{eq:baseline-cancels}
V^{\pi_\theta}(c_t) \sum_{v \in \mathcal{V}} \pi_\theta(v\mid c_t)\,\nabla_\theta\log \pi_\theta(v\mid c_t)
\;=\; V^{\pi_\theta}(c_t)\,\nabla_\theta \!\sum_{v \in \mathcal{V}} \pi_\theta(v\mid c_t) \;=\; 0.
\end{equation}
Because $\text{OPD}_\text{full-V}$ computes the full KL, its gradient already has zero variance at $c_t$, leaving nothing for the baseline to reduce. The baseline becomes useful only once we replace the full-vocabulary expectation with a Monte Carlo estimate, as is done in \method$_\text{full-V}$.

\subsection{Top-$k$ Approximation}
\label{sec:vopd-topk}
While \method$_\text{full-V}$ adds no additional backward-pass cost, it still requires the exact KL computation at $O(|\mathcal{V}|)$ cost per token. Similar to $\text{OPD}_\text{top-$k$}$ (Eq.~\eqref{eq:opd-topk-J}), we can approximate the baseline KL using only the student's top-$k$ support to further reduce compute:
\begin{equation}
\label{eq:vopd-topk-baseline}
\hat b_t(c_t)
\;=\;
-\,\mathbb{D}_\mathrm{KL}\!\bigl(\bar\pi_\theta(\cdot\mid c_t) \,\big\|\, \bar\pi_T(\cdot\mid c_t)\bigr),
\end{equation}
where $\bar\pi$ is the renormalized distribution on the student's top-$k$ support $S_t$ with $k \ll |\mathcal{V}|$. Substituting $\hat b_t$ into Eq.~\eqref{eq:baseline-grad} gives the \method$_\text{top-$k$}$ gradient estimator:
\begin{equation}
\label{eq:vopd-topk-grad}
\nabla_\theta \mathcal{J}_{\text{\method}_\text{top-$k$}}(\theta)
\;=\;
\mathbb{E}\!\left[\, \sum_{t=1}^{|y|}
\underbrace{\Bigl( r_t(c_t,y_t) + \mathbb{D}_\mathrm{KL}\!\bigl(\bar\pi_\theta(\cdot\mid c_t) \,\big\|\, \bar\pi_T(\cdot\mid c_t)\bigr) \Bigr)}_{\textstyle a_t(c_t,y_t)}
\,\nabla_\theta \log \pi_\theta(y_t\mid c_t) \,\right].
\end{equation}

\paragraph{The crucial distinction from $\text{OPD}_\text{top-$k$}$.}
While both methods compute KL with a top-$k$ approximation, they place it differently in the estimator. $\text{OPD}_\text{top-$k$}$ uses it as the \emph{loss}, replacing $\mathbb{D}_\mathrm{KL}(\pi_\theta\|\pi_T)$ with $\mathbb{D}_\mathrm{KL}(\bar\pi_\theta\|\bar\pi_T)$, thus changing the optimization target and biasing the gradient. \method$_\text{top-$k$}$ uses it as a \emph{detached baseline} subtracted from the reward. As discussed in \S~\ref{sec:baseline}, because $\hat b_t$ depends only on $\pi_\theta$, $\pi_T$, and $S_t$ but not on the sampled token $y_t$, the gradient remains unbiased. Furthermore, since $\hat b_t$ still approximates the value function $V^{\pi_\theta}(c_t)$, it can reduce variance. The same approximation in different positions thus has completely different consequences, which we further confirm empirically in \S~\ref{sec:exp-math}: \method$_\text{top-$k$}$ yields substantial gains over OPD in practice, while $\text{OPD}_\text{top-$k$}$ largely does not.

\paragraph{Robustness to approximation error.}
Unbiasedness holds for any $k$, but does variance reduction require an accurate baseline? We show that it does not, as even a coarse top-$k$ estimate retains most of the benefit. Let $\epsilon_t=\mathbb{D}_\mathrm{KL}(\pi_\theta\|\pi_T)-\mathbb{D}_\mathrm{KL}(\bar\pi_\theta\|\bar\pi_T)$ denote the approximation error of $\hat b_t$, and $R(b)$ the variance reduction achieved by a baseline $b$. Under the same approximation used in Eq.~\eqref{eq:variance-reduction},
\begin{equation}
\label{eq:topk-variance}
\frac{R(\hat b_t)}{R(V^{\pi_\theta})}
\;\approx\;
1-\left(\frac{\epsilon_t}{\mathbb{D}_\mathrm{KL}(\pi_\theta\|\pi_T)}\right)^2.
\end{equation}
Two consequences follow. First, the loss in variance reduction is quadratic in the \emph{relative} approximation error and independent of its sign, so a coarse estimate suffices. For example, a $10\%$ relative error still retains approximately $99\%$ of the reduction, which helps explain why even $k{=}5$, despite visible approximation error, performs comparably to \method$_\text{full-V}$ (see \S~\ref{sec:exp_ablation}). Second, the approximate baseline still reduces variance whenever $|\epsilon_t|<\mathbb{D}_\mathrm{KL}(\pi_\theta\|\pi_T)$, for which a sufficient condition is that the student’s top-$k$ support contains over half of its probability mass. We derive these results in \S~\ref{app:topk-variance}.

\begin{table}[t]
\centering
\caption{Comparison of OPD variants along key algorithmic axes. Our methods are highlighted.}
\label{tab:method-axes}
\renewcommand{\arraystretch}{1.25}
\begin{tabular}{lccccc}
\toprule
\textbf{Method}
  & \makecell{\textbf{Gradient} \\ \textbf{unbiased?}}
  & \makecell{\textbf{Gradient} \\ \textbf{variance at $\mathbf{c_t}$}}
  & \makecell{\textbf{Backward} \\ \textbf{token count}}
  & \makecell{\textbf{Per-token} \\ \textbf{KL cost}}
  & \makecell{\textbf{Total additional} \\ \textbf{compute}} \\
\midrule
OPD                                         & \cmark & High & 1                 & ---                & None      \\
$\text{OPD}_\text{full-V}$                  & \cmark & None & $|\mathcal{V}|$   & $O(|\mathcal{V}|)$ & High      \\
$\text{OPD}_\text{top-$k$}$                 & \xmark & None & $k$               & $O(k)$             & Medium    \\
\rowcolor{blue!10} \method$_\text{full-V}$  & \cmark & Low  & 1                 & $O(|\mathcal{V}|)$ & Low       \\
\rowcolor{blue!10} \method$_\text{top-$k$}$ & \cmark & Low  & 1                 & $O(k)$             & Very Low  \\
\bottomrule
\end{tabular}
\end{table}

\subsection{Summary: Algorithm Comparison}
\label{sec:opd-comparison}
Table~\ref{tab:method-axes} compares the discussed algorithms along the key axes of bias, variance, and compute. Base OPD is the computationally lightest variant but suffers from high gradient variance due to its single-sample Monte Carlo estimator. $\text{OPD}_\text{full-V}$ eliminates this sampling variance by computing the per-token KL at $c_t$ over the full vocabulary, but requires $O(|\mathcal{V}|)$ cost for both the per-token KL computation and the backward pass. \text{OPD}$_\text{top-$k$}$ reduces both costs to $O(k)$, but changes the original objective by restricting the KL to a truncated support, thereby biasing the gradient. In contrast, \method$_\text{full-V}$ preserves base OPD’s unbiased single-token estimator while reducing variance via the value baseline, adding only a per-token KL computation in the forward pass. \method{}$_\text{top-$k$}$ further approximates this baseline on the student’s top-$k$ support, preserving unbiasedness while achieving variance reduction at the lowest additional compute. Overall, among all compared methods, only \method{} combines unbiasedness, reduced variance, and a single-token backward pass.

\section{Experiments}
\subsection{Experimental Setup}
\label{sec:exp_setup}
\paragraph{Models and methods.} Our primary setting distills Qwen3-1.7B into Qwen3-1.7B-Base~\citep{yang2025qwen3}, mirroring a common industrial OPD configuration in which a post-trained checkpoint is distilled back into its base model~\citep{zeng2026glm, yang2026nemotron2, deepseekai2026deepseekv4}. We additionally vary three axes: (i) scale, with Qwen3-4B into Qwen3-4B-Base and Qwen3-14B into Qwen3-14B-Base; (ii) teacher--student capability gap, with Qwen3-1.7B and Qwen3-14B into Qwen3-0.6B-Base; and (iii) model family, with Olmo-3-7B-Think into Olmo-3-7B-Base~\citep{olmo2025olmo} and Gemma-4-12B-it into Gemma-4-E2B-it~\citep{team2026gemma}. We evaluate both \method{}$_\text{full-V}$ and \method{}$_\text{top-$k$}$ against the three OPD variants from \S~\ref{sec:opd} (base OPD, $\text{OPD}_\text{full-V}$, and $\text{OPD}_\text{top-$k$}$). For $\text{OPD}_\text{top-$k$}$, we set $k{=}20$ following~\citet{li2026rethinkingopd}, who show that gains saturate beyond $k{=}16$. For \method{}$_\text{top-$k$}$, we likewise default to $k{=}20$ and verify robustness in \S~\ref{sec:exp_ablation}.

\paragraph{Mathematical reasoning.} Following prior OPD work~\citep{lu2025onpolicydistillation, li2026rethinkingopd}, we use mathematical reasoning as our primary domain and evaluate all seven teacher--student pairs in this setting. We train on the 14K-sample English subset of DAPO-Math-17K~\citep{yu2025dapo}, a standard RLVR training set, for a single epoch. We evaluate on MATH500~\citep{hendrycks2021math, lightman2024let}, Minerva Math~\citep{lewkowycz2022solving}, AMC23~\citep{maa_amc}, and AIME24/25~\citep{maa_aime}, reporting avg@$n$ and pass@$n$ with $n{=}8$ for MATH500 and Minerva Math and $n{=}32$ for the smaller AMC23 and AIME24/25 benchmarks. We use maximum response lengths of 2048 and 4096 tokens and sampling temperatures of 1.0 and 0.6 for training and evaluation, respectively. All runs use LoRA~\citep{hu2022lora} for parameter-efficient training. See \S~\ref{app:exp_settings} for detailed settings.

\paragraph{Additional domains.} To test generalization, we apply the primary Qwen3-1.7B setting to scientific reasoning and instruction following. For science, we use the chemistry subset of SciKnowEval~\citep{feng2024sciknoweval}, partitioned into train/validation/test splits of 75/5/20 following recent practice~\citep{kim2026does, hubotter2026sdpo, zhang2026illusion, shenfeld2026selfdistill}, and evaluate on its test set and GPQA-Diamond~\citep{rein2024gpqa}. For instruction following, we train on RLVR-IFeval~\citep{lambert2024tulu}\footnote{\url{https://huggingface.co/datasets/allenai/RLVR-IFeval}} and evaluate on IFEval~\citep{zhou2023instruction}. We report avg@8 on all three benchmarks.

\begin{table}[!t]
\centering
\caption{Performance on mathematical reasoning benchmarks. Best performance is \textbf{bolded}, and second best performance is \underline{underlined}.}
\label{tab:math-results}
\resizebox{\textwidth}{!}{%
\begin{tabular}{l|cc|cc|cc|cc|c}
\toprule
& \multicolumn{2}{c|}{\textbf{MATH500}} & \multicolumn{2}{c|}{\textbf{MINERVA}} & \multicolumn{2}{c|}{\textbf{AMC23}} & \multicolumn{2}{c|}{\textbf{AIME24/25}} & \\
\cmidrule(lr){2-3} \cmidrule(lr){4-5} \cmidrule(lr){6-7} \cmidrule(lr){8-9}
\textbf{Method} & Avg@8 & Pass@8 & Avg@8 & Pass@8 & Avg@32 & Pass@32 & Avg@32 & Pass@32 & \textbf{Avg.} \\
\midrule
\multicolumn{10}{c}{\textit{Qwen3-1.7B $\rightarrow$ Qwen3-1.7B-Base}} \\
\midrule
Student & 42.3 & 79.8 & 13.8 & 37.1 & 23.2 & \underline{85.0} & 3.0 & 20.0 & 20.6 \\
OPD & 58.7 & 82.6 & 22.2 & 40.8 & 33.4 & 77.5 & 4.8 & \underline{28.3} & 29.8 \\
OPD$_{\text{top-}k}$ & 58.0 & 84.0 & 23.4 & \underline{44.5} & 35.5 & \underline{85.0} & 4.0 & \textbf{30.0} & 30.2 \\
OPD$_{\text{full-V}}$ & \underline{64.6} & \textbf{85.4} & 25.0 & \underline{44.5} & \textbf{36.5} & \textbf{87.5} & \underline{5.8} & 25.0 & \underline{33.0} \\
\rowcolor{blue!10}
\method{}$_{\text{top-}k}$ & \textbf{64.9} & \underline{84.8} & \underline{25.2} & \underline{44.5} & \underline{36.1} & 82.5 & 5.6 & 26.7 & \underline{33.0} \\
\rowcolor{blue!10}
\method{}$_{\text{full-V}}$ & 64.0 & 84.6 & \textbf{26.2} & \textbf{48.9} & \underline{36.1} & 80.0 & \textbf{6.2} & 25.0 & \textbf{33.1} \\
\midrule
\multicolumn{10}{c}{\textit{Qwen3-4B $\rightarrow$ Qwen3-4B-Base}} \\
\midrule
Student & 51.1 & 86.4 & 15.9 & 45.2 & 38.7 & 90.0 & 7.8 & 26.8 & 28.4 \\
OPD & 75.2 & 90.8 & 35.3 & \underline{54.0} & 50.5 & 90.0 & 10.4 & 38.3 & 42.9 \\
OPD$_{\text{top-}k}$ & 75.0 & 91.1 & 35.3 & 52.2 & 50.7 & \textbf{95.0} & 10.1 & \textbf{41.7} & 42.8 \\
OPD$_{\text{full-V}}$ & 78.6 & 91.8 & 36.3 & 53.3 & 51.0 & 87.5 & \textbf{14.3} & \underline{40.0} & 45.1 \\
\rowcolor{blue!10}
\method{}$_{\text{top-}k}$ & \textbf{79.3} & \textbf{93.0} & \textbf{37.6} & \textbf{54.4} & \underline{51.2} & 87.5 & 13.2 & \textbf{41.7} & \underline{45.3} \\
\rowcolor{blue!10}
\method{}$_{\text{full-V}}$ & \underline{78.9} & \underline{92.4} & \underline{37.2} & \underline{54.0} & \textbf{52.0} & \underline{92.5} & \underline{13.5} & \underline{40.0} & \textbf{45.4} \\
\midrule
\multicolumn{10}{c}{\textit{Olmo-3-7B-Think $\rightarrow$ Olmo-3-7B-Base}} \\
\midrule
Student & 43.1 & \underline{83.4} & 11.6 & 35.3 & 20.6 & \textbf{70.0} & 5.4 & \textbf{30.0} & 20.2 \\
OPD & 61.2 & 78.6 & 18.4 & 36.0 & 32.5 & 50.0 & 7.4 & 21.7 & 29.9 \\
OPD$_{\text{top-}k}$ & 58.8 & 75.4 & 21.5 & 39.0 & 32.3 & 45.0 & 5.5 & 15.0 & 29.5 \\
OPD$_{\text{full-V}}$ & 62.8 & \textbf{83.6} & 21.7 & \textbf{43.0} & 34.0 & 62.5 & \textbf{8.6} & 23.4 & 31.8 \\
\rowcolor{blue!10}
\method{}$_{\text{top-}k}$ & \underline{64.0} & 81.0 & \textbf{23.9} & \underline{41.9} & \underline{35.9} & 57.5 & \underline{8.4} & \underline{25.0} & \textbf{33.1} \\
\rowcolor{blue!10}
\method{}$_{\text{full-V}}$ & \textbf{64.4} & 83.2 & \underline{22.5} & 40.4 & \textbf{36.8} & \underline{65.0} & 7.9 & 21.7 & \underline{32.9} \\
\bottomrule
\end{tabular}%
}
\end{table}

\subsection{Mathematical Reasoning Results}
\label{sec:exp-math}
Table~\ref{tab:math-results} summarizes our primary results on Qwen3-1.7B, Qwen3-4B, and Olmo-3-7B. Across all three configurations, \method{} consistently improves over base OPD by a substantial margin. In the Qwen3-1.7B-Base setting, \method$_\text{top-$k$}$ and \method$_\text{full-V}$ achieve absolute gains of up to +6.2\% on MATH500 and above +3\% on average. These improvements extend to the 4B scale, where both \method{} variants gain around +4\% on MATH500 and around +2.5\% on average over base OPD, and to the Olmo-3-7B family, where \method$_\text{top-$k$}$ reaches an average of 33.1\% compared to 29.9\% for OPD. On the most challenging AIME24/25, \method{} improves avg@32 in all three configurations, most notably from 10.4\% to 13.5\% at 4B. Crucially, across all settings the two \method{} variants, \method$_\text{full-V}$ and \method$_\text{top-$k$}$, achieve nearly identical performance, confirming that the top-$k$ baseline approximation captures the essential variance reduction without loss of accuracy. In contrast, OPD$_\text{top-$k$}$ yields only marginal changes over base OPD, for example +0.4\% average at 1.7B, consistent with the finding of~\citet{li2026rethinkingopd} and potentially attributable to the bias in its objective. Overall, \method{} performs competitively with, and sometimes exceeds, OPD$_\text{full-V}$, which requires a full-vocabulary backward pass at every token, while adding only a lightweight forward-pass computation.

Table~\ref{tab:appendix-math-results} in \S~\ref{app:extended-results} extends these results to four additional teacher--student pairs, namely Qwen3-14B into Qwen3-14B-Base for scale, Gemma-4-12B-it into Gemma-4-E2B-it for model family, and Qwen3-1.7B and Qwen3-14B into Qwen3-0.6B-Base for capability gap. \method{} improves over base OPD in every setting, by up to +3.6\% on average, and matches or exceeds OPD$_\text{full-V}$ nearly everywhere at substantially lower cost, while OPD$_\text{top-$k$}$ trails. The contrast is largest under a severe capability gap, where distilling Qwen3-14B into Qwen3-0.6B-Base with base OPD degrades the student's average from 12.3\% to 11.6\%, whereas \method$_\text{full-V}$ improves it to 13.8\% (+7.4\% over OPD on MATH500). These results suggest that the control variate baseline is a general and robust stabilizer for OPD, especially when the teacher and student distributions differ substantially. The appendix additionally shows that base OPD can collapse during training while \method{} remains stable, that \method{} outperforms the clipping method REOPOLD~\citep{ko2026scaling}, and that its gains hold across maximum response lengths.

\subsection{Further Analysis}
\begin{figure}[!ht]
    \centering
    \includegraphics[width=\linewidth]{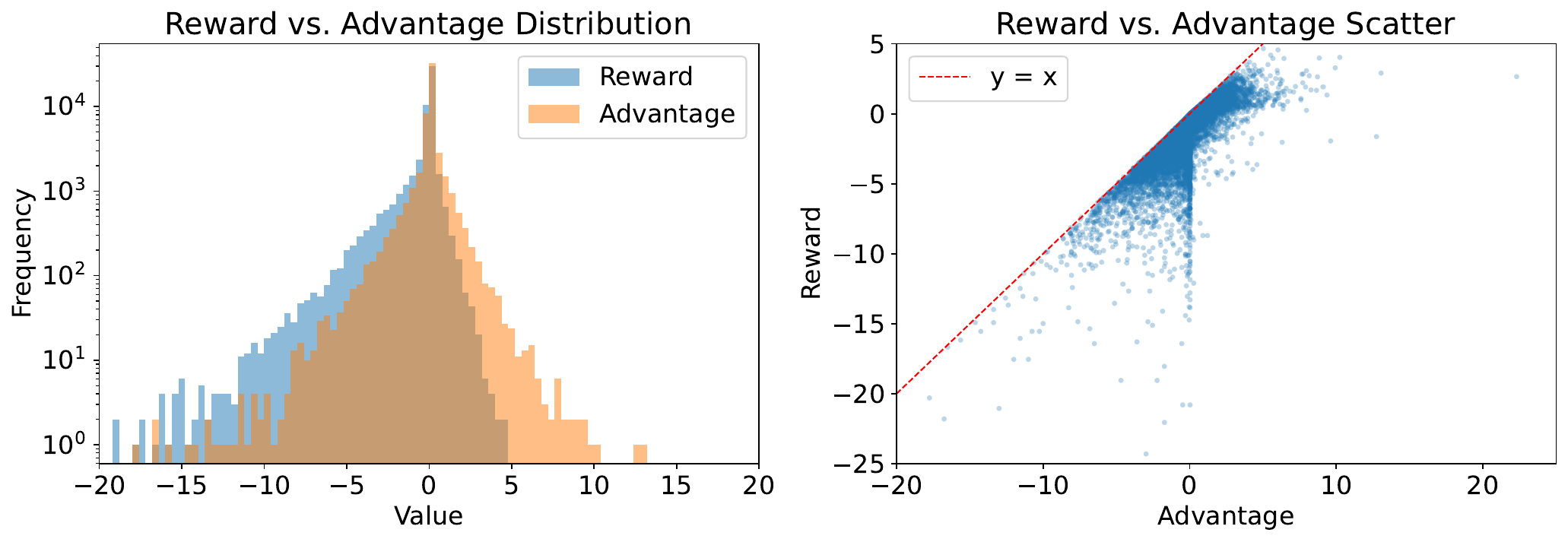} 
    \caption{Token-level reward and advantage distributions. \textbf{Left:} The marginal distributions. \textbf{Right:} Per-token scatter plot (x: advantage, y: reward).}
    \label{fig:rewards}
\end{figure}

\paragraph{Advantage vs.\ Reward.} To understand how \method{} reshapes the training signal, we compare the per-token reward $r_t$ with the advantage $a_t = r_t + \mathbb{D}_\mathrm{KL}(\pi_\theta\|\pi_T)$, logging all token-level values from the first training batch of 64 prompts (approximately 55k tokens) in the Qwen3-1.7B into Qwen3-1.7B-Base setting. As shown in Figure~\ref{fig:rewards} (left), the OPD reward distribution exhibits a pronounced negative long tail driven by high-mismatch tokens, consistent with recent reports~\citep{ko2026scaling} and our discussion in \S~\ref{sec:vopd_value}. In contrast, the \method{} advantage distribution is shifted rightward with this tail compressed toward zero, as expected since $\mathbb{D}_\mathrm{KL}(\pi_\theta\|\pi_T) \geq 0$ makes the shift always non-negative.

\label{sec:exp-further}

Figure~\ref{fig:rewards} (right) plots the per-token advantage (x-axis) against reward (y-axis). All points lie on or to the right of $y = x$, confirming that the baseline only shifts rewards positively. Notably, positive-reward tokens are largely unchanged, whereas among tokens with similarly negative rewards, some are dampened almost entirely to zero while others retain advantages close to their rewards. Because the baseline is the KL divergence at context $c_t$, tokens at high-KL contexts receive a large positive shift that absorbs most of the negative reward, while those at low-KL contexts are left nearly intact.

This selectivity has a natural interpretation. A large negative reward arises when the student assigns high probability to a token the teacher considers unlikely, and suppressing this token likely shifts its mass toward the student's other high-probability candidates. In low-KL contexts, the teacher likely also assigns high probability to these alternatives, so the gradient is informative and \method{} has minimal influence. In high-KL contexts, however, these alternatives may be less probable under the teacher, yielding a harmful gradient that can be mitigated by the large baseline of \method{}. This is consistent with Eq.~\eqref{eq:variance-reduction}, where variance reduction scales with $\mathbb{D}_\mathrm{KL}(\pi_\theta\|\pi_T)^2$ and is thus largest exactly where updates are least informative. Prior work has identified these high-mismatch tokens as the dominant source of gradient instability in OPD~\citep{li2026rethinkingopd}, and the fact that \method{}'s selective suppression improves rather than degrades accuracy (Table~\ref{tab:math-results}) confirms that what is removed is noise rather than signal---an effect that simple reward clipping cannot replicate (see \S~\ref{app:extended-results}).

\begin{figure}[t]
    \centering
    \includegraphics[width=\linewidth]{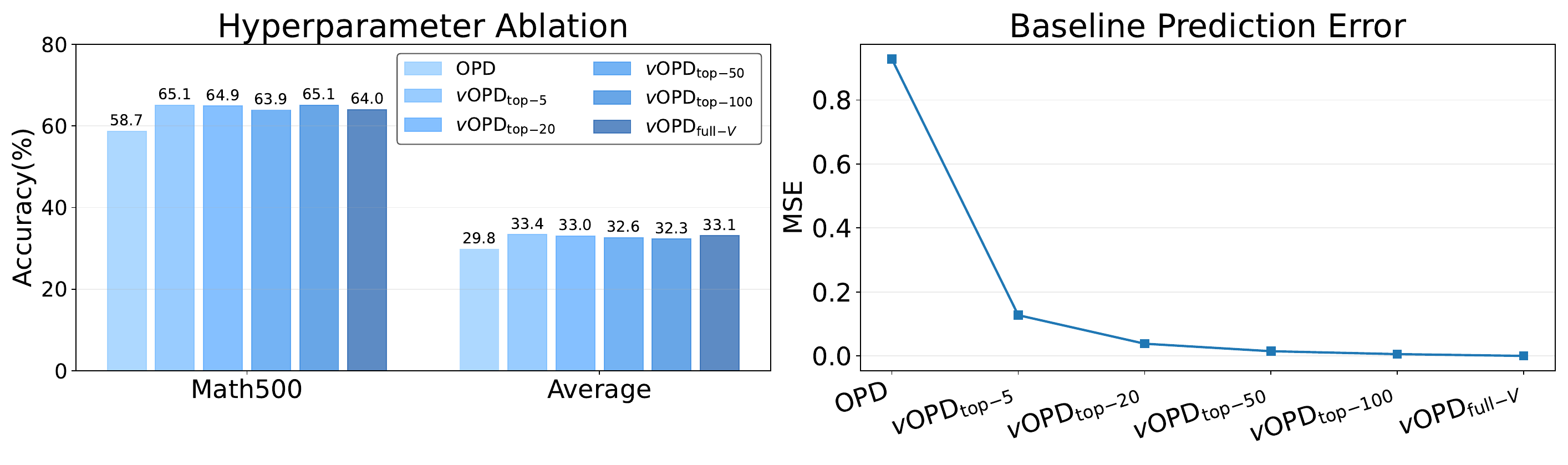} 
    \caption{\textbf{Left:} Average accuracy for various $k$ on \method{}$_\text{top-$k$}$. \textbf{Right:} Mean squared error of the top-$k$ KL baseline relative to the full-vocabulary KL.}
    \label{fig:ablation}
    \vspace{-1.0em}
\end{figure}

\paragraph{Hyperparameter Sensitivity.}
\label{sec:exp_ablation}
We ablate $k$ in \method$_\text{top-$k$}$ in the Qwen3-1.7B into Qwen3-1.7B-Base setting. Figure~\ref{fig:ablation} (left) shows that average accuracy is stable across $k \in \{5, 20, 50, 100\}$ and \method$_\text{full-V}$, all substantially outperforming OPD, which corresponds to $k{=}0$ (no baseline). Any nonzero $k$ thus suffices, as even the coarsest approximation at $k{=}5$ provides enough variance reduction to stabilize training (see Table~\ref{tab:appendix-math-results}, bottom). Figure~\ref{fig:ablation} (right) plots the mean squared error of the top-$k$ KL estimate relative to the full-vocabulary KL; for base OPD, this equals the squared KL itself, since no baseline is subtracted. The error decreases monotonically with $k$ and approaches zero beyond $k{=}20$. Notably, $k{=}5$ carries non-trivial approximation error yet still matches the accuracy of \method$_\text{full-V}$, suggesting that a coarse value estimate suffices for stable training, consistent with our analysis in \S~\ref{sec:vopd-topk}.

\paragraph{Wall-Clock Time.}
\begin{figure}
    \centering
    \vspace{-1.0em}
    \includegraphics[width=\linewidth]{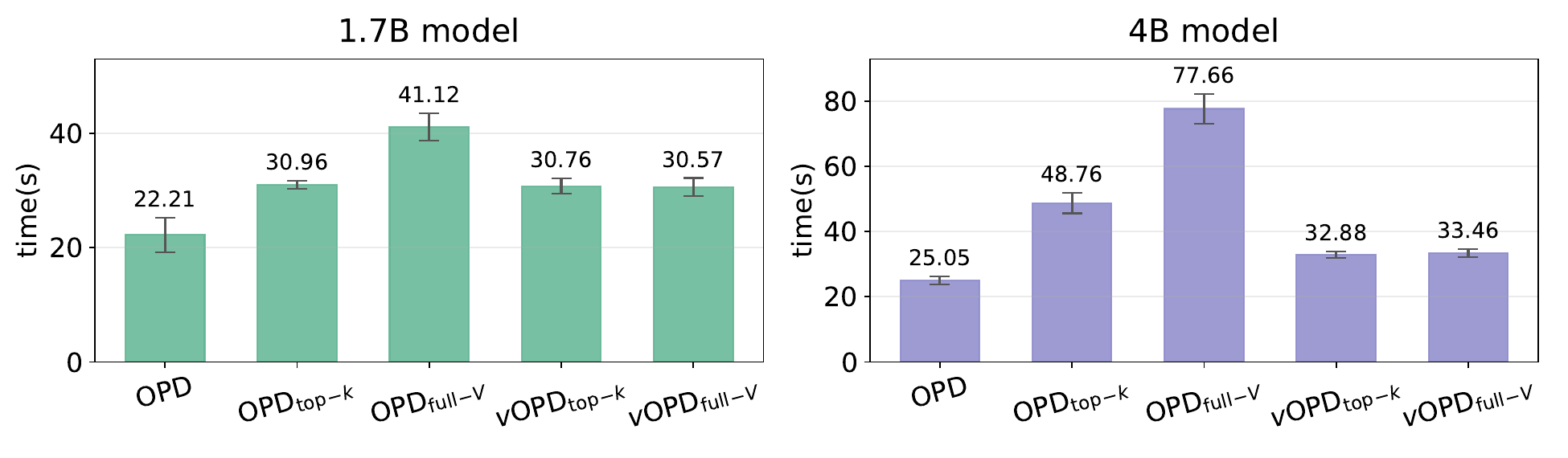}
    \caption{Per-step wall-clock time for OPD variants at 1.7B and 4B scale.  The error bars denote variance.}
    \label{fig:time}
    \vspace{-1.0em}
\end{figure}
Figure~\ref{fig:time} compares per-step wall-clock time across all five methods at the Qwen3-1.7B and 4B scales on a single NVIDIA H200 GPU. At 1.7B, base OPD is the fastest, followed by \method$_\text{top-$k$}$, \method$_\text{full-V}$, and OPD$_\text{top-$k$}$, which cluster at a modest overhead, while OPD$_\text{full-V}$ is the most expensive due to its full-vocabulary backward pass. At 4B, the gaps widen. \method{} reduces per-step time by up to 57.7\% relative to OPD$_\text{full-V}$, OPD$_\text{top-$k$}$ becomes considerably more expensive than both \method{} variants, and \method$_\text{top-$k$}$ pulls slightly ahead of \method$_\text{full-V}$, while the overall ordering remains the same. Combined with Table~\ref{tab:math-results}, \method$_\text{top-$k$}$ offers the best accuracy-to-compute tradeoff among all compared methods, and the widening gap at 4B highlights the scalability advantage of \method{}.
 
\begin{figure}[t]
    \centering
    \includegraphics[width=\linewidth]{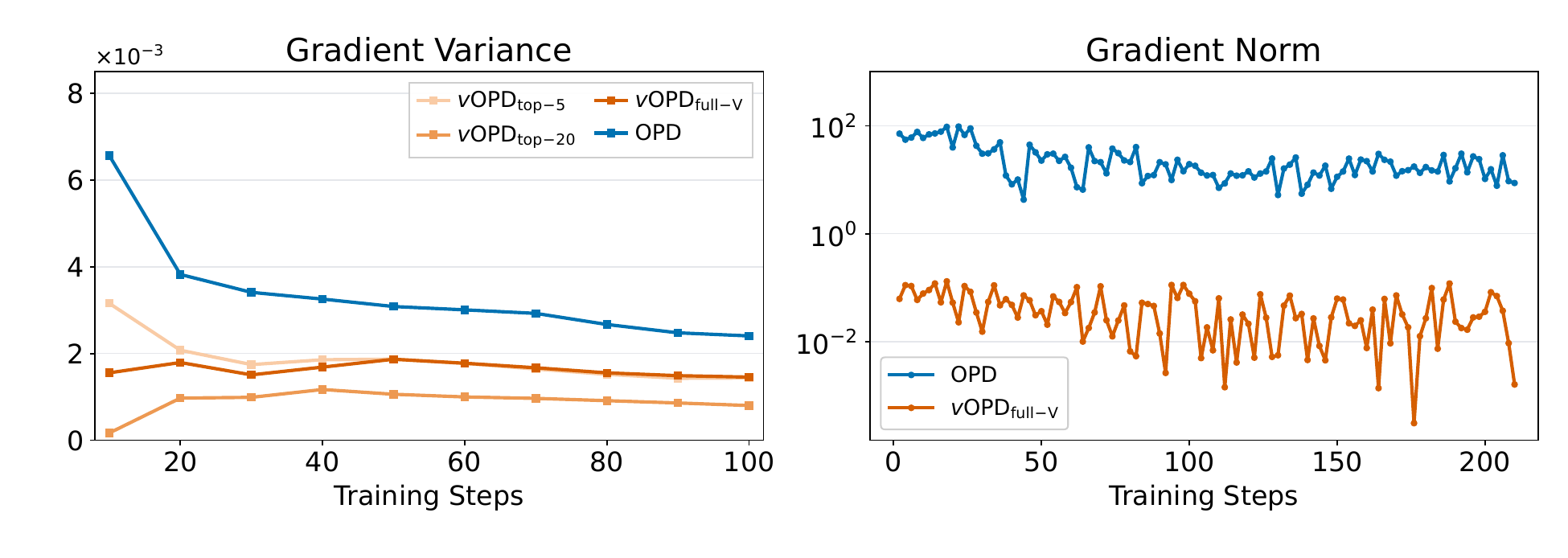}
    \caption{Gradient statistics during training in the Qwen3-1.7B setting. \textbf{Left:} Trace of the gradient variance. \textbf{Right:} Gradient norm.}
    \label{fig:grad_stats}
\end{figure}

\paragraph{Gradient Variance and Norm.}
We directly measure the variance reduction predicted by Eq.~\eqref{eq:variance-reduction} in the Qwen3-1.7B setting, estimating $\mathrm{tr}(\mathrm{Var}[g]) = \sum_i (\mathbb{E}[g_i^2] - \mathbb{E}[g_i]^2)$ from AdamW's moment EMAs~\citep{takashiro2026advantage}, which adds no training overhead (Figure~\ref{fig:grad_stats}, left). All \method{} variants reduce run-averaged variance by 45--74\% relative to base OPD. The gap is largest early in training, where the student and teacher disagree most, consistent with the $\mathbb{D}_\mathrm{KL}^2$ scaling in Eq.~\eqref{eq:variance-reduction}. The reduction also holds for the top-$k$ baselines, with top-20 even falling below the exact value baseline. This is possible because the value function is itself only an approximation of the variance-optimal baseline (see \S~\ref{app:variance}), so an approximate baseline can land closer to the optimum.

Gradient norms are likewise lower throughout training (Figure~\ref{fig:grad_stats}, right), with \method$_\text{full-V}$ keeping norms orders of magnitude below base OPD. Despite these smaller gradients, \method{} trains stably and reaches higher accuracy (Table~\ref{tab:math-results}), suggesting that the large gradients in base OPD are dominated by variance rather than useful learning signal. Together, these results provide direct empirical evidence that \method{} stabilizes OPD by reducing gradient variance.

\begin{wraptable}{r}{0.45\textwidth}
\vspace{-1.25em}
\centering
\caption{Performance beyond mathematical reasoning in the Qwen3-1.7B setting. Best performance is \textbf{bolded}, and second best is \underline{underlined}.}
\label{tab:science-ablation}
\small
\setlength{\tabcolsep}{4pt}
\begin{tabular}{l|cc|c}
\toprule
\textbf{Method} & SciKnowEval & GPQA-D & IFEval \\
\midrule
Student
& 26.1 & 20.4 & 19.8 \\
OPD
& 29.3 & 24.7 & 30.6 \\
OPD$_{\text{top-}k}$
& 29.7 & 24.1 & 34.0 \\
OPD$_{\text{full-V}}$
& \textbf{35.1} & \textbf{28.7} & \textbf{37.7} \\
\rowcolor{blue!10}
\method{}$_{\text{top-}k}$
& 33.2 & \underline{28.6} & 34.8 \\
\rowcolor{blue!10}
\method{}$_{\text{full-V}}$
& \underline{34.7} & 28.4 & \underline{36.4} \\
\bottomrule
\end{tabular}
\vspace{-1.0em}
\end{wraptable}
\subsection{Additional Domains Results}
\label{sec:exp-sci}
Table~\ref{tab:science-ablation} shows that \method{} generalizes beyond mathematical reasoning. On SciKnowEval and GPQA-Diamond, both \method{} variants improve over base OPD by roughly +4--5\%, while OPD$_\text{top-$k$}$ provides little improvement. On instruction following, \method{} likewise improves over base OPD on IFEval by up to +5.8\%. Here OPD$_\text{top-$k$}$ is more competitive, gaining +3.4\%, but still trails \method$_\text{full-V}$ by 2.4\%. Although OPD$_\text{full-V}$ attains the highest scores in these domains, \method$_\text{full-V}$ remains within 1.3\% of it on every benchmark at substantially lower cost. These results support the generality of the control variate mechanism across various domains.

\section{Related Work}
\subsection{On-Policy Distillation}
\label{sec:related-opd}
OPD has become an important component of LLM post-training, especially for long Chain-of-Thought reasoning tasks~\citep{wei2022chain}, where dense token-level teacher signals offer a compute-efficient alternative to sparse RLVR rewards. Early works such as GKD~\citep{agarwal2024onpolicy} and MiniLLM~\citep{gu2024minillm} established OPD as an effective alternative to standard distillation. Recent work has popularized token-level Monte Carlo OPD~\citep{lu2025onpolicydistillation}, studied practical recipes~\citep{patino2025opd,li2026rethinkingopd}, and incorporated OPD into large-scale post-training systems~\citep{yang2025qwen3,zeng2026glm,yang2026nemotron2,deepseekai2026deepseekv4}. Despite these successes, OPD remains unstable in practice. Follow-up studies have explored top-$k$ support restrictions~\citep{li2026rethinkingopd}, clipping~\citep{ko2026scaling}, entropy-aware training~\citep{jin2026entropy}, and prefix-only variants~\citep{zhang2026prefixopd}. Our work complements these efforts by treating OPD instability as an estimator-variance problem and addressing it with an unbiased control variate baseline.

\subsection{Control Variate Baseline for RL}
\label{sec:related-baselines}
The control variate baseline is a core tool in on-policy policy-gradient reinforcement learning. This principle underlies actor-critic methods, advantage estimation, and modern policy-gradient algorithms such as A3C and PPO with GAE~\citep{williams1992simple,sutton1999policygrad,mnih2016asynchronous,schulman2017ppo,schulman2016high}. The same idea remains central in LLM reinforcement learning. Early RLHF pipelines used PPO~\citep{ouyang2022rlhf, schulman2017ppo} with a learned value model to estimate advantages, while recent reasoning-oriented RLVR methods~\citep{lambert2024tulu} often replace the learned critic with a simpler relative baseline. GRPO and RLOO, for example, construct a relative baseline from the rewards of multiple sampled responses and have become standard recipes for RLVR~\citep{ahmadian2024back,shao2024deepseekmath,guo2025deepseek, yu2025dapo}. Follow-up methods such as SPO explore alternative single-stream baseline estimators~\citep{xu2025spo}, and POISE integrates mechanistic interpretability techniques for baseline estimation~\citep{choi2026your}. Despite the central role of the baseline in RL, it has not been systematically explored for OPD, even though OPD admits a policy-gradient interpretation~\citep{lu2025onpolicydistillation,ko2026scaling,jin2026entropy}. Our work fills this gap by deriving a closed-form OPD value baseline and using it as an unbiased control variate.

\section{Conclusion, Limitations, and Future Work}
\label{sec:conclusion}
We introduced \method{}, a control variate formulation of On-Policy Distillation that reduces the variance of the single-sample Monte Carlo estimator without changing the original OPD objective. By using the negative reverse KL between the student and the teacher as a detached value baseline, \method{} preserves an unbiased policy-gradient estimator while retaining the single-token backward pass of base OPD. Across mathematics, science, and instruction following, \method{} consistently reduces gradient variance and improves stability and performance over base OPD, matching full-vocabulary OPD at lower cost and acting as a principled regularizer on destabilizing negative reward tokens.

Several directions remain open for future work. Our experiments are limited to models up to 14B parameters, and validating \method{} at a larger scale is a natural next step. Wall-clock results reflect our implementation and are not definitive, and further optimizing \method$_\text{top-$k$}$ could widen its speed advantage over \method$_\text{full-V}$. As a distillation method, \method{} requires access to a stronger teacher, and extending it to self-distillation settings is a possible direction. Finally, this work focuses on the token-level KL in OPD, and considering sequence-level KL objectives is another extension.

\section*{Acknowledgment}
This work was supported by the National Research Foundation of Korea (NRF) grant RS-2024-00333484 and the Institute of Information \& Communications Technology Planning \& Evaluation (IITP) grant under the AI Computing Support Project for R\&D (RS-2026-25505492, High-Performance Research AI Computing Infrastructure Support at the 2 PFLOPS Scale), both funded by the Korean government (MSIT). This research was also supported by the Korea Institute of Science and Technology Information (KISTI) in 2026 (No. (KISTI)K26L3M1C1) as part of the development of KONI (KISTI Open Neural Intelligence), a large language model specialized in science and technology. The National Supercomputing Center provided supercomputing resources and technical support (KSC-2025-CRE-0514).

\newpage
\bibliographystyle{plainnat}
\bibliography{neurips}
\newpage
\appendix

\section{Theoretical Derivations}
\label{app:theory}

This appendix provides derivations omitted from \S~\ref{sec:prelim} and \S~\ref{sec:method}. Following the main sections, we denote the context $c_t=(x,y_{<t})$, and the reward $r_t(c_t,y_t) = \log\pi_T(y_t\mid c_t) - \log\pi_\theta(y_t\mid c_t)$. We use the score-function identity:
\begin{equation}
\label{eq:app-score}
\sum_{v} \pi_\theta(v\mid c_t)\,\nabla_\theta\log\pi_\theta(v\mid c_t)
\;=\; \sum_{v} \nabla_\theta\,\pi_\theta(v\mid c_t)
\;=\; \nabla_\theta \sum_{v} \pi_\theta(v\mid c_t)
\;=\; 0,
\end{equation}
which follows from the log trick $\nabla_\theta\log\pi_\theta(v\mid c_t) = \nabla_\theta\pi_\theta(v\mid c_t)/\pi_\theta(v\mid c_t)$.

\subsection{Unbiasedness of Baseline Subtraction}
\label{app:unbiased}

We show that subtracting an action-independent baseline $b_t(c_t)$ from the per-token reward leaves the policy gradient unbiased~\citep{williams1992simple,sutton1999policygrad}.
Specifically, we show:
\begin{equation}
\label{eq:app-unbiased-claim}
\mathbb{E}_{y_t \sim \pi_\theta(\cdot\mid c_t)}\!\Bigl[
  \bigl(r_t(y_t) - b_t(c_t)\bigr)\,\nabla_\theta\log\pi_\theta(y_t\mid c_t)
\Bigr]
\;=\;
\mathbb{E}_{y_t \sim \pi_\theta(\cdot\mid c_t)}\!\Bigl[
  r_t(y_t)\,\nabla_\theta\log\pi_\theta(y_t\mid c_t)
\Bigr].
\end{equation}

By linearity of expectation, the difference between the two sides equals
\begin{equation}
    \mathbb{E}_{y_t \sim \pi_\theta(\cdot\mid c_t)}\!\Bigl[
  b_t(c_t)\,\nabla_\theta\log\pi_\theta(y_t\mid c_t)
\Bigr]
\;=\;
b_t(c_t)
\sum_{v} \pi_\theta(v\mid c_t)\,\nabla_\theta\log\pi_\theta(v\mid c_t),
\end{equation}
where $b_t(c_t)$ factors out of the expectation because it is independent of $y_t$ by assumption. The score-function identity (Eq.~\eqref{eq:app-score}) gives the remaining sum as zero, so the difference vanishes and Eq.~\eqref{eq:app-unbiased-claim} holds.

\subsection{Variance Reduction and the Optimal Baseline}
\label{app:variance}

We derive the variance-reducing property of baseline subtraction and identify the optimal scalar choice~\citep{greensmith2004variance}. Consider the per-step gradient estimator:
\begin{equation}
    g(b) \;=\; \bigl(r_t(y_t) - b\bigr)\,\nabla_\theta\log\pi_\theta(y_t\mid c_t),
\qquad b \in \mathbb{R}.
\end{equation}
We seek the $b$ minimizing $\mathrm{tr}(\mathrm{Var}[g(b)])$. $\mathbb{E}[g(b)]$ does not depend on $b$ because baseline subtraction is unbiased (see \S~\ref{app:unbiased}), so the variance is minimized by minimizing the second moment $\mathbb{E}[\|g(b)\|^2]$. Expanding,
\begin{align}
\mathbb{E}\!\bigl[\|g(b)\|^2\bigr]
&\;=\;
  \mathbb{E}\!\Bigl[
    \bigl(r_t(y_t) - b\bigr)^2\,
    \bigl\|\nabla_\theta\log\pi_\theta(y_t\mid c_t)\bigr\|^2
  \Bigr] \\[4pt]
&\;=\;
  \mathbb{E}\!\Bigl[r_t(y_t)^2\,\bigl\|\nabla_\theta\log\pi_\theta(y_t\mid c_t)\bigr\|^2\Bigr]
  \;-\; 2b\,\mathbb{E}\!\Bigl[r_t(y_t)\,\bigl\|\nabla_\theta\log\pi_\theta(y_t\mid c_t)\bigr\|^2\Bigr] \\
&\quad+\; b^2\,\mathbb{E}\!\Bigl[\bigl\|\nabla_\theta\log\pi_\theta(y_t\mid c_t)\bigr\|^2\Bigr].
\end{align}

This is a convex quadratic in $b$. Setting its derivative to zero yields the
\emph{optimal scalar baseline} $b^\star$:
\begin{equation}
\label{eq:app-optimal-baseline}
b^\star
\;=\;
\frac{
  \mathbb{E}_{y_t \sim \pi_\theta(\cdot\mid c_t)}\!\Bigl[
    r_t(y_t)\,\bigl\|\nabla_\theta\log\pi_\theta(y_t\mid c_t)\bigr\|^2
  \Bigr]
}{
  \mathbb{E}_{y_t \sim \pi_\theta(\cdot\mid c_t)}\!\Bigl[
    \bigl\|\nabla_\theta\log\pi_\theta(y_t\mid c_t)\bigr\|^2
  \Bigr]
}.
\end{equation}
The optimal baseline in Eq.~\eqref{eq:app-optimal-baseline} is a weighted form of
the value function. In practice, the value function itself is used as the baseline,
motivating the choice in \S~\ref{sec:vopd_value}:
\begin{equation}
    b^\star
\;\approx\;
\mathbb{E}_{y_t \sim \pi_\theta(\cdot\mid c_t)}\!\bigl[r_t(y_t)\bigr]
\;=\; V^{\pi_\theta}(c_t).
\end{equation}

\subsection{Variance Reduction of \texorpdfstring{\method{}}{vOPD}}
\label{app:variance-derivation}
We now derive Eq.~\eqref{eq:variance-reduction}, the per-step variance reduction
obtained by \method$_\text{full-V}$. Recall the two estimators:
\begin{equation}
g_\text{OPD} = r_t(c_t, y_t)\,\nabla_\theta\log\pi_\theta(y_t\mid c_t),
\qquad
g_{\text{\method}_{\text{full-V}}} = \bigl(r_t(c_t, y_t) - b_t\bigr)\,\nabla_\theta\log\pi_\theta(y_t\mid c_t),
\end{equation}
where $b_t = V^{\pi_\theta}(c_t) = -\mathbb{D}_\mathrm{KL}(\pi_\theta(\cdot\mid c_t)\,\|\,\pi_T(\cdot\mid c_t))$. We want to show:
\begin{equation}
\label{eq:app-vopd-variance-claim}
\mathrm{tr}\!\bigl(\mathrm{Var}[g_\text{OPD}]\bigr) - \mathrm{tr}\!\bigl(\mathrm{Var}[g_{\text{\method}_{\text{full-V}}}]\bigr)
\;\approx\;
\mathbb{D}_\mathrm{KL}\!\bigl(\pi_\theta(\cdot\mid c_t)\,\big\|\,\pi_T(\cdot\mid c_t)\bigr)^{2}
\;\cdot\;
\mathbb{E}\!\bigl[\|\nabla_\theta\log\pi_\theta(y_t\mid c_t)\|^2\bigr].
\end{equation}
We start from the second-moment expansion of \S~\ref{app:variance}. Since both
estimators share the same expectation (see \S~\ref{app:unbiased}), the variance
difference equals the difference of second moments:
\begin{align*}
&\mathrm{tr}\!\bigl(\mathrm{Var}[g_\text{OPD}]\bigr) - \mathrm{tr}\!\bigl(\mathrm{Var}[g_{\text{\method}_{\text{full-V}}}]\bigr) \\
&\quad= \mathbb{E}\!\bigl[r_t^2\,\|\nabla_\theta\log\pi_\theta(y_t\mid c_t)\|^2\bigr]
- \mathbb{E}\!\bigl[(r_t - b_t)^2\,\|\nabla_\theta\log\pi_\theta(y_t\mid c_t)\|^2\bigr] \\
&\quad= 2b_t\,\mathbb{E}\!\bigl[r_t\,\|\nabla_\theta\log\pi_\theta(y_t\mid c_t)\|^2\bigr]
- b_t^2\,\mathbb{E}\!\bigl[\|\nabla_\theta\log\pi_\theta(y_t\mid c_t)\|^2\bigr].
\end{align*}
Under a weak-correlation approximation $\mathbb{E}[r_t\,\|\nabla_\theta\log\pi_\theta(y_t\mid c_t)\|^2] \approx b_t\,\mathbb{E}[\|\nabla_\theta\log\pi_\theta(y_t\mid c_t)\|^2]$, this yields:
\begin{equation}
\mathrm{tr}\!\bigl(\mathrm{Var}[g_\text{OPD}]\bigr) - \mathrm{tr}\!\bigl(\mathrm{Var}[g_{\text{\method}_{\text{full-V}}}]\bigr)
\;\approx\;
b_t^2\,\mathbb{E}\!\bigl[\|\nabla_\theta\log\pi_\theta(y_t\mid c_t)\|^2\bigr].
\end{equation}
Substituting $b_t^2 = \mathbb{D}_\mathrm{KL}(\pi_\theta\|\pi_T)^2$ recovers Eq.~\eqref{eq:app-vopd-variance-claim}.

\begin{figure}[!th]
    \centering
    \includegraphics[width=\linewidth]{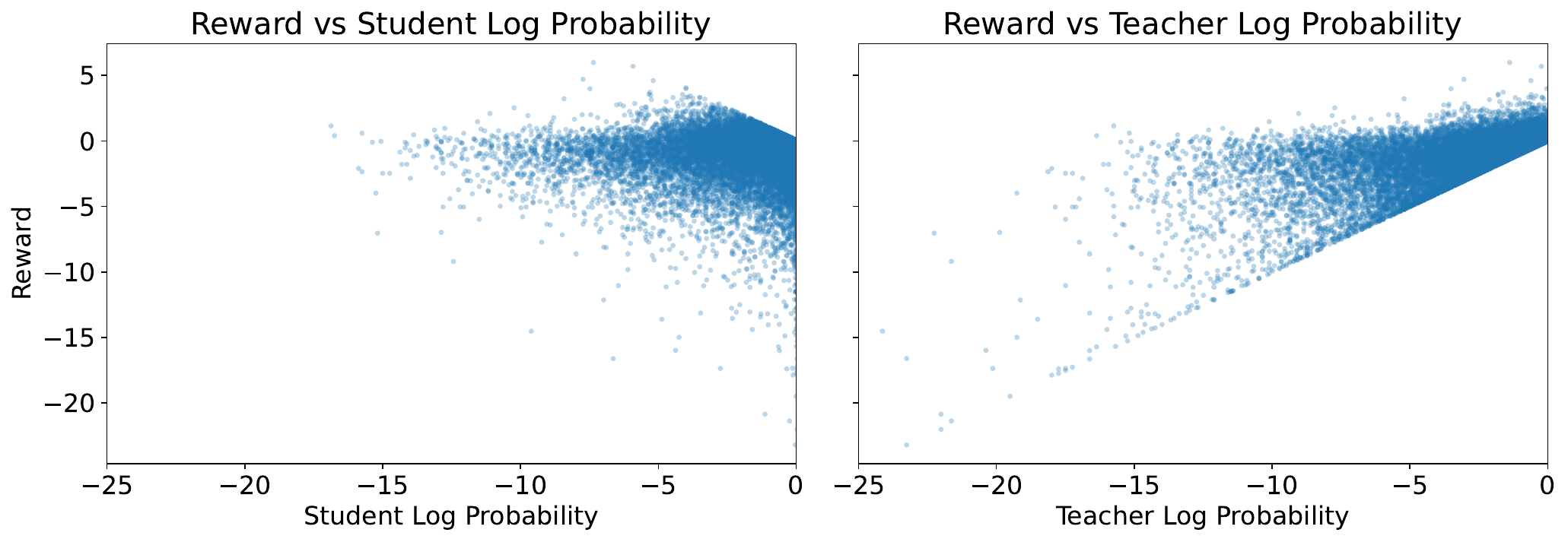}
    \caption{Scatter plot. \textbf{Left:} student log probability ($x$) vs.\ reward ($y$).
    \textbf{Right:} teacher log probability ($x$) vs.\ reward ($y$).}
    \label{fig:logp-scatter}
\end{figure}

\paragraph{Weak-correlation approximation.}
We discuss why a weak-correlation assumption is reasonable in the OPD setting. Because $y_t$ is sampled from $\pi_\theta$, sampled tokens naturally lie in the high-probability region, so $\log\pi_\theta(y_t\mid c_t)$ varies in a narrow band and
$\|\nabla_\theta\log\pi_\theta(y_t\mid c_t)\|^2$, a function of $\pi_\theta$ at $y_t$, can inherit this concentration. In contrast, the variation in $r_t = \log\pi_T(y_t\mid c_t) - \log\pi_\theta(y_t\mid c_t)$ is primarily driven by the teacher term, which has no structural dependence on $\pi_\theta(y_t \mid c_t)$. Figure~\ref{fig:logp-scatter} confirms this empirically, showing scatter plots from the same data used in \S~\ref{sec:exp-further}: $\log\pi_\theta(y_t\mid c_t)$ concentrates near zero with weak correlation to $r_t$, while $\log\pi_T(y_t \mid c_t)$ spans a wider range and exhibits a clear correlation with $r_t$. This is also consistent with the analysis in \S~\ref{sec:exp-further} and recent long-tail negative reward structure report~\citep{ko2026scaling}: since $\log\pi_\theta(y_t \mid c_t)$ stays close to zero, $r_t$ cannot spike in the positive direction, but can extend into large negative values whenever $\log\pi_T(y_t \mid c_t)$ drops---making the teacher term the dominant driver of variation in $r_t$.

\subsection{Variance Reduction under the Top-\texorpdfstring{$k$}{k} Baseline}
\label{app:topk-variance}
Fix a context $c_t$. For an action-independent baseline $b$, let $g_t(b)=(r_t-b)\nabla_\theta\log\pi_\theta(y_t\mid c_t)$, and define its variance reduction relative to no baseline as:
\begin{equation}
\label{eq:app-R-def}
R(b)
=
\mathrm{tr}\bigl(\mathrm{Var}[g_t(0)]\bigr)
-
\mathrm{tr}\bigl(\mathrm{Var}[g_t(b)]\bigr).
\end{equation}
Because baseline subtraction leaves the expected gradient unchanged, $R(b)$ equals the difference in second moments:
\begin{equation}
\label{eq:app-R-approx}
\begin{split}
R(b)
&=
2b\,\mathbb{E}\bigl[r_t\|\nabla_\theta\log\pi_\theta(y_t\mid c_t)\|^2\bigr]-b^2\,\mathbb{E}\bigl[\|\nabla_\theta\log\pi_\theta(y_t\mid c_t)\|^2\bigr] \\
&\approx
\bigl(2b\,V^{\pi_\theta}(c_t)-b^2\bigr)\,\mathbb{E}\bigl[\|\nabla_\theta\log\pi_\theta(y_t\mid c_t)\|^2\bigr],
\end{split}
\end{equation}
where the approximation uses the same weak-correlation assumption as \S~\ref{app:variance-derivation}. Under this approximation, the value baseline $V^{\pi_\theta}(c_t)=-\mathbb{D}_\mathrm{KL}(\pi_\theta\|\pi_T)$ maximizes the variance reduction, recovering Eq.~\eqref{eq:variance-reduction}.

For the top-$k$ baseline $\hat b_t=-\mathbb{D}_\mathrm{KL}(\bar\pi_\theta\|\bar\pi_T)$, let $\epsilon_t=\mathbb{D}_\mathrm{KL}(\pi_\theta\|\pi_T)-\mathbb{D}_\mathrm{KL}(\bar\pi_\theta\|\bar\pi_T)$ as in \S~\ref{sec:vopd-topk}. Since $\hat b_t=V^{\pi_\theta}(c_t)+\epsilon_t$, substituting into Eq.~\eqref{eq:app-R-approx} gives, for $\mathbb{D}_\mathrm{KL}(\pi_\theta\|\pi_T)>0$,
\begin{equation}
\label{eq:app-topk-R}
\begin{split}
R(\hat b_t)
&\approx
\bigl(\mathbb{D}_\mathrm{KL}(\pi_\theta\|\pi_T)^2-\epsilon_t^2\bigr)\,\mathbb{E}\bigl[\|\nabla_\theta\log\pi_\theta(y_t\mid c_t)\|^2\bigr], \\
\frac{R(\hat b_t)}{R(V^{\pi_\theta})}
&\approx
1-\left(\frac{\epsilon_t}{\mathbb{D}_\mathrm{KL}(\pi_\theta\|\pi_T)}\right)^2.
\end{split}
\end{equation}

\paragraph{Quadratic tolerance to approximation error.}
The loss in variance reduction is quadratic in the relative error $\epsilon_t/\mathbb{D}_\mathrm{KL}(\pi_\theta\|\pi_T)$. Thus, a $10\%$ relative error retains approximately $99\%$ of the reduction, while a $30\%$ error retains approximately $91\%$. The sign of $\epsilon_t$ is irrelevant, consistent with the empirical robustness across $k$ in \S~\ref{sec:exp_ablation}.

\paragraph{Sufficient condition for variance reduction.}
From Eq.~\eqref{eq:app-topk-R}, the approximate baseline reduces variance whenever
\begin{equation}
\label{eq:app-topk-condition}
|\epsilon_t|<\mathbb{D}_\mathrm{KL}(\pi_\theta\|\pi_T)
\qquad\Longleftrightarrow\qquad
0<\mathbb{D}_\mathrm{KL}(\bar\pi_\theta\|\bar\pi_T)<2\,\mathbb{D}_\mathrm{KL}(\pi_\theta\|\pi_T).
\end{equation}
The KL chain rule over the partition $\{S_t,S_t^c\}$ gives
\begin{equation}
\mathbb{D}_\mathrm{KL}(\pi_\theta\|\pi_T)\;\geq\;\pi_\theta(S_t\mid c_t)\,\mathbb{D}_\mathrm{KL}(\bar\pi_\theta\|\bar\pi_T),
\end{equation}
so $\pi_\theta(S_t\mid c_t)>\nicefrac{1}{2}$ implies the upper bound in Eq.~\eqref{eq:app-topk-condition}. Hence, whenever the student's top-$k$ support contains more than half of its probability mass, the top-$k$ baseline never increases variance, and strictly reduces it whenever $\mathbb{D}_\mathrm{KL}(\bar\pi_\theta\|\bar\pi_T)>0$.

\section{Experiment Settings}
\label{app:exp_settings}
This section details the experimental settings from \S~\ref{sec:exp_setup}. All experiments were conducted on a single NVIDIA H200 NVL or A100 NVL GPU, paired with an Intel(R) Xeon(R) Gold 6530 or Gold 6230 CPU @ 2.10GHz, respectively. All training runs took between 2h (Qwen 1.7B on A100) and 6h (Olmo 7B on H200).

\subsection{Training Settings}
Table~\ref{tab:hyperparameters_all} summarizes the hyperparameters for training and evaluation.
Following recent work demonstrating that the maximum response length in OPD need not be excessively long (2--3K tokens suffice), we set the maximum response length to 2048~\citep{zhang2026prefixopd, li2026rethinkingopd}. Following recent work on OPD, we set the student sampling temperature to 1.0~\citep{li2026rethinkingopd,ko2026scaling}. We adopt parameter-efficient training via LoRA~\citep{hu2022lora}.
Our implementation builds on \texttt{TRL}'s MiniLLM\footnote{\url{https://huggingface.co/docs/trl/main/minillm}} implementation~\citep{vonwerra2020trl,gu2024minillm}. When using Qwen as the teacher model, we disable its extended thinking mode. For scientific reasoning, which comprises 1,890 train samples from the chemistry subset of SciKnowEval~\citep{feng2024sciknoweval}, we extend training to a maximum of 10 epochs.

For evaluation, following the official guidelines for Qwen3 and Olmo-3 \cite{yang2025qwen3, olmo2025olmo}, we set the sampling temperature to 0.6 and top-p sampling to 0.9. We use \texttt{vLLM} for accelerated inference~\citep{kwon2023efficient}.

\begin{table}[!htb]
\centering
\caption{Training and Evaluation Settings}
\label{tab:hyperparameters_all}
\small
\setlength{\tabcolsep}{4pt}
\begin{tabular}{l|c|c||l|c|c}
\toprule
\textbf{Hyperparameter} & \textbf{Train} & \textbf{Eval}
& \textbf{Hyperparameter} & \textbf{Train} & \textbf{Eval} \\
\midrule
Prompt length   & 1024 & 1024
& Response length & 2048 & 4096 \\
Rollout temperature     & 1.0 & 0.6
& Top-$p$ sampling        & 1.0 & 0.9 \\
Batch size              & 64 & --
& Optimizer              & AdamW & -- \\
Learning rate           & 1e-5 (Qwen3), 2e-5 (others) & --
& LoRA rank               & 64 & -- \\
Epochs                  & 10 (science), 1 (others) & --
& LoRA $\alpha$          & 128 & -- \\
Engine                  & \texttt{TRL} & \texttt{vLLM}
& Precision               & \texttt{bfloat16} & \texttt{bfloat16} \\
\bottomrule
\end{tabular}
\end{table}

\subsection{Prompts}
We use the following prompts for mathematical and scientific reasoning throughout training and evaluation, following the official prompts~\citep{yang2025qwen3}.
\begin{tcolorbox}[
    colback=gray!5,
    colframe=gray!50,
    title=\textbf{Math Reasoning Prompt Template},
    fonttitle=\bfseries,
    boxrule=0.5pt,
    arc=2pt,
]
\small
\texttt{\{problem\}}\textbackslash n\textbackslash nPlease reason step by step, and put your final answer within \textbackslash \textbackslash boxed\{\}.

\vspace{0.5em}

\end{tcolorbox}

\begin{tcolorbox}[
    colback=gray!5,
    colframe=gray!50,
    title=\textbf{Scientific Reasoning Prompt Template},
    fonttitle=\bfseries,
    boxrule=0.5pt,
    arc=2pt,
]
\small
\texttt{\{problem\}}\textbackslash n\textbackslash nReturn only one JSON object with the key "answer". The value must be exactly one choice letter, for example: \{"answer": "C"\}.
\end{tcolorbox}

\begin{table}[!ht]
\centering
\caption{Extended mathematical reasoning results. Avg.\ is the mean of the Avg@$n$ columns. Best performance within each block is \textbf{bolded}, and second best is \underline{underlined}.}
\label{tab:appendix-math-results}
\resizebox{\textwidth}{!}{%
\begin{tabular}{l|cc|cc|cc|cc|c}
\toprule
& \multicolumn{2}{c|}{\textbf{MATH500}}
& \multicolumn{2}{c|}{\textbf{MINERVA}}
& \multicolumn{2}{c|}{\textbf{AMC23}}
& \multicolumn{2}{c|}{\textbf{AIME24/25}}
& \\
\cmidrule(lr){2-3}
\cmidrule(lr){4-5}
\cmidrule(lr){6-7}
\cmidrule(lr){8-9}
\textbf{Method}
& Avg@8 & Pass@8
& Avg@8 & Pass@8
& Avg@32 & Pass@32
& Avg@32 & Pass@32
& \textbf{Avg.} \\
\midrule
\multicolumn{10}{c}{\textit{Teacher Performance}} \\
\midrule
Qwen3-1.7B
& 73.0 & 90.8 & 28.6 & 42.6 & 44.8 & 85.0 & 14.4 & 35.0 & 40.2 \\
Qwen3-4B
& 83.5 & 93.8 & 43.9 & 49.6 & 65.9 & 95.0 & 20.2 & 51.7 & 53.4 \\
Olmo-3-7B-Think
& 87.2 & 95.4 & 39.2 & 52.2 & 70.0 & 97.5 & 37.2 & 68.3 & 58.4 \\
\midrule
\multicolumn{10}{c}{\textit{Qwen3-14B $\rightarrow$ Qwen3-14B-Base}} \\
\midrule
Student
& 78.5 & 92.2 & 35.2 & 53.7 & \underline{60.4} & \textbf{95.0} & 13.1 & 43.3 & 46.8 \\
OPD
& 81.3 & 92.0 & 34.8 & 53.8 & 58.8 & 90.0 & 19.0 & \textbf{53.4} & 48.4 \\
OPD$_{\text{top-}k}$
& 83.4 & \underline{94.0} & \textbf{39.4} & \underline{55.1} & 59.4 & \underline{92.5} & 19.5 & 48.4 & 50.4 \\
OPD$_{\text{full-V}}$
& 83.3 & 93.6 & \underline{39.0} & 54.0 & 58.8 & \underline{92.5} & 18.7 & 45.0 & 50.0 \\
\rowcolor{blue!10}
\method{}$_{\text{top-}k}$
& \underline{85.0} & \textbf{94.4} & 38.7 & \textbf{55.2} & 59.2 & 90.0 & \textbf{20.5} & 46.0 & \underline{50.8} \\
\rowcolor{blue!10}
\method{}$_{\text{full-V}}$
& \textbf{85.2} & \textbf{94.4} & 35.9 & 51.5 & \textbf{62.1} & \textbf{95.0} & \underline{20.4} & \underline{51.7} & \textbf{50.9} \\
\midrule
\multicolumn{10}{c}{\textit{Gemma-4-12B-it $\rightarrow$ Gemma-4-E2B-it}} \\
\midrule
Student
& 76.3 & \underline{93.6} & 36.4 & 49.6 & 67.1 & 95.0 & 28.2 & 55.0 & 52.0 \\
OPD
& 76.9 & \underline{93.6} & 36.5 & 49.6 & 69.2 & 95.0 & 28.8 & 55.0 & 52.9 \\
OPD$_{\text{top-}k}$
& \underline{78.9} & \textbf{93.8} & \underline{37.4} & 48.2 & 70.0 & \textbf{97.5} & 28.8 & 60.0 & 53.8 \\
OPD$_{\text{full-V}}$
& \textbf{80.3} & 92.6 & \textbf{37.8} & \underline{50.7} & 71.3 & \textbf{97.5} & \textbf{33.7} & 55.0 & \textbf{55.7} \\
\rowcolor{blue!10}
\method{}$_{\text{top-}k}$
& 76.7 & \textbf{93.8} & 35.3 & 48.2 & \underline{72.8} & 95.0 & \underline{32.9} & \textbf{65.0} & 54.4 \\
\rowcolor{blue!10}
\method{}$_{\text{full-V}}$
& 78.3 & 93.0 & \underline{37.4} & \textbf{51.1} & \textbf{73.5} & \textbf{97.5} & 31.7 & \underline{63.3} & \underline{55.2} \\
\midrule
\multicolumn{10}{c}{\textit{Qwen3-1.7B $\rightarrow$ Qwen3-0.6B-Base}} \\
\midrule
Student
& 30.9 & 66.8 & 5.6 & 21.7 & 12.3 & 60.0 & 0.4 & 6.7 & 12.3 \\
OPD
& 41.4 & \underline{73.0} & 8.1 & 26.1 & 19.3 & \textbf{80.0} & 1.1 & \underline{13.3} & 17.5 \\
OPD$_{\text{top-}k}$
& 43.7 & 72.0 & 12.7 & \textbf{32.4} & \underline{21.8} & 65.0 & \textbf{1.5} & \textbf{13.4} & 19.9 \\
OPD$_{\text{full-V}}$
& \textbf{47.3} & 72.0 & 12.8 & 29.4 & 20.7 & 65.0 & \underline{1.4} & \underline{13.3} & \underline{20.6} \\
\rowcolor{blue!10}
\method{}$_{\text{top-}k}$
& 44.4 & 71.0 & \underline{13.7} & 29.0 & 20.9 & \underline{67.5} & 0.9 & 10.0 & 20.0 \\
\rowcolor{blue!10}
\method{}$_{\text{full-V}}$
& \underline{46.5} & \textbf{73.2} & \textbf{14.3} & \underline{32.0} & \textbf{22.3} & \underline{67.5} & \underline{1.4} & 10.0 & \textbf{21.1} \\
\midrule
\multicolumn{10}{c}{\textit{Qwen3-14B $\rightarrow$ Qwen3-0.6B-Base (Severe Capability Gap)}} \\
\midrule
Student
& \underline{30.9} & \textbf{66.8} & 5.6 & 21.7 & \textbf{12.3} & \textbf{60.0} & \textbf{0.4} & \textbf{6.7} & \underline{12.3} \\
OPD
& 28.3 & 56.6 & \underline{8.8} & \textbf{27.2} & 9.3 & 40.0 & 0.1 & \underline{1.7} & 11.6 \\
\rowcolor{blue!10}
\method{}$_{\text{full-V}}$
& \textbf{35.7} & \underline{62.0} & \textbf{9.8} & \textbf{27.2} & \underline{9.7} & \underline{47.5} & \underline{0.2} & \underline{1.7} & \textbf{13.8} \\
\midrule
\multicolumn{10}{c}{\textit{Qwen3-1.7B $\rightarrow$ Qwen3-1.7B-Base (Comparison with REOPOLD)}} \\
\midrule
OPD
& 58.7 & 82.6 & \underline{22.2} & 40.8 & \underline{33.4} & 77.5 & 4.8 & \textbf{28.3} & 29.8 \\
REOPOLD ($\lambda{=}0.3$)
& \underline{62.1} & 80.8 & 20.6 & 41.2 & 32.3 & 72.5 & 5.0 & \underline{26.7} & \underline{30.0} \\
REOPOLD ($\lambda{=}0.1$)
& 61.8 & \textbf{85.8} & 19.2 & \underline{44.1} & 32.3 & \textbf{82.5} & \textbf{6.6} & 16.7 & \underline{30.0} \\
\rowcolor{blue!10}
\method{}$_{\text{full-V}}$
& \textbf{64.0} & \underline{84.6} & \textbf{26.2} & \textbf{48.9} & \textbf{36.1} & \underline{80.0} & \underline{6.2} & 25.0 & \textbf{33.1} \\
\midrule
\multicolumn{10}{c}{\textit{Qwen3-1.7B $\rightarrow$ Qwen3-1.7B-Base (Top-$k$ Ablation)}} \\
\midrule
OPD
& 58.7 & 82.6 & 22.2 & 40.8 & 33.4 & 77.5 & 4.8 & \textbf{28.3} & 29.8 \\
\method{}$_{\text{top-}5}$
& \textbf{65.1} & 83.8 & \underline{26.0} & \underline{45.2} & \underline{35.6} & \underline{85.0} & \textbf{6.7} & \underline{26.7} & \textbf{33.4} \\
\method{}$_{\text{top-}20}$
& \underline{64.9} & \textbf{84.8} & 25.2 & 44.5 & \textbf{36.1} & 82.5 & 5.6 & \underline{26.7} & 33.0 \\
\method{}$_{\text{top-}50}$
& 63.9 & \underline{84.6} & 25.6 & 44.1 & 35.0 & \textbf{87.5} & 6.1 & 23.3 & 32.6 \\
\method{}$_{\text{top-}100}$
& \textbf{65.1} & 76.4 & 25.7 & 32.0 & 33.4 & 75.0 & 5.2 & \underline{26.7} & 32.3 \\
\method{}$_{\text{full-V}}$
& 64.0 & \underline{84.6} & \textbf{26.2} & \textbf{48.9} & \textbf{36.1} & 80.0 & \underline{6.2} & 25.0 & \underline{33.1} \\
\bottomrule
\end{tabular}%
}
\end{table}

\section{Extended Experiment Results}
\label{app:extended-results}
Table~\ref{tab:appendix-math-results} presents the extended mathematical reasoning results. Specifically, it reports (i) the accuracy of the teacher models; (ii) the four additional teacher--student pairs from \S~\ref{sec:exp-math}, covering scale (Qwen3-14B into Qwen3-14B-Base), model family (Gemma-4-12B-it into Gemma-4-E2B-it), and capability gap (Qwen3-1.7B and Qwen3-14B into Qwen3-0.6B-Base); (iii) a comparison with the clipping-based method REOPOLD~\citep{ko2026scaling}; and (iv) full results for the top-$k$ ablation in Figure~\ref{fig:ablation}.

\paragraph{Comparison with reward clipping.} REOPOLD~\citep{ko2026scaling} stabilizes OPD by clipping the per-token reward with a uniform threshold. We evaluate it in the Qwen3-1.7B setting with two $\lambda$ thresholds, including the recommended value. \method{} outperforms both by over +3\% on average, indicating that the context-adaptive KL baseline is more effective than uniform clipping.

\paragraph{Training stability.} Table~\ref{tab:stability} tracks MATH500 Avg@8 during a representative Qwen3-4B into Qwen3-4B-Base run under the configuration of \S~\ref{sec:exp_setup}. Base OPD improves quickly but then collapses below its starting accuracy, while \method$_\text{full-V}$ improves steadily and remains stable. This suggests that the accuracy reachable by base OPD is limited by its stability, so the accuracy gains and the stabilization of \method{} are not independent effects.

\begin{table}[h]
\centering
\caption{MATH500 Avg@8 over training steps (Qwen3-4B $\rightarrow$ Qwen3-4B-Base).}
\label{tab:stability}
\begin{tabular}{lccccc}
\toprule
Method & Step 0 & Step 25 & Step 50 & Step 75 & Step 100 \\
\midrule
OPD & 51.1 & 75.2 & 69.4 & 36.4 & 25.8 \\
\rowcolor{blue!10}
\method$_\text{full-V}$ & 51.1 & 76.3 & 78.9 & 78.6 & 78.8 \\
\bottomrule
\end{tabular}
\end{table}

\paragraph{Response length.} We vary the maximum response length in the Qwen3-1.7B setting (Table~\ref{tab:length}). \method{} outperforms base OPD at every length, including 512 tokens, a quarter of the default budget. Performance degrades only mildly at shorter lengths, indicating that the benefit of \method{} is not tied to a particular response length, consistent with recent findings~\citep{zhang2026prefixopd, li2026rethinkingopd}. Since shorter budgets proportionally reduce cost, this also offers a practical accuracy--compute tradeoff.

\begin{table}[h]
\centering
\caption{Avg@$n$ across maximum response lengths (Qwen3-1.7B $\rightarrow$ Qwen3-1.7B-Base).}
\label{tab:length}
\begin{tabular}{llccccc}
\toprule
MaxLen & Method & MATH500 & Minerva & AMC23 & AIME24/25 & Avg. \\
\midrule
512 & OPD & 56.8 & 18.0 & 30.2 & 3.3 & 27.1 \\
\rowcolor{blue!10}
 & \method$_\text{full-V}$ & 63.1 & 21.2 & 35.5 & 5.5 & 31.3 \\
\midrule
1024 & OPD & 58.8 & 20.0 & 30.3 & 3.7 & 28.2 \\
\rowcolor{blue!10}
 & \method$_\text{full-V}$ & 62.9 & 21.3 & 35.8 & 6.3 & 31.5 \\
\midrule
2048 & OPD & 58.7 & 22.2 & 33.4 & 4.8 & 29.8 \\
\rowcolor{blue!10}
 & \method$_\text{full-V}$ & 64.0 & 26.2 & 36.1 & 6.2 & 33.1 \\
\bottomrule
\end{tabular}
\end{table}

\clearpage

\end{document}